\documentclass[preprint,11pt]{elsarticle}

\usepackage{amssymb}

\usepackage{fullpage}
\usepackage{url}
\usepackage{verbatim}
\usepackage{wrapfig}

\journal{Neuroimage Clinical}

\begin{document}

\begin{frontmatter}



\title{A multi-path 2.5 dimensional convolutional neural network system for segmenting stroke lesions in brain MRI images}


\author[1]{Yunzhe Xue}

\author[1]{Fadi G. Farhat}

\author[2,3]{Olga Boukrina}

\author[2,3]{A. M. Barrett}

\author[4]{Jeffrey R. Binder}

\author[1]{Usman W. Roshan\corref{cor1}}
\ead{usman@njit.edu}

\author[5]{William W. Graves}

\cortext[cor1]{Corresponding author}

\address[1]{Department of Computer Science, New Jersey Institute of Technology, Newark, NJ  07102, USA}

\address[2]{Stroke Rehabilitation Research, Kessler Foundation, West Orange, NJ, USA}

\address[3]{Department of Physical Medicine and Rehabilitation, Rutgers -- New Jersey Medical School, Newark, NJ, USA}

\address[4]{Department of Neurology, Medical College of Wisconsin, Milwaukee, WI, USA}

\address[5]{Department of Psychology, Rutgers University -- Newark, Newark, NJ, USA}


\begin{abstract}
Automatic identification of brain lesions from magnetic resonance imaging (MRI) scans of stroke survivors would be a useful aid in patient diagnosis and treatment planning. It would also greatly facilitate the study of brain-behavior relationships by eliminating the laborious step of having a human expert manually segment the lesion on each brain scan. We propose a multi-modal multi-path convolutional neural network system for automating stroke lesion segmentation. Our system has nine end-to-end UNets that take as input 2-dimensional (2D) slices and examines all three planes with three different normalizations. Outputs from these nine total paths are concatenated into a 3D volume that is then passed to a 3D convolutional neural network to output a final lesion mask. We trained and tested our method on datasets from three sources: Medical College of Wisconsin (MCW), Kessler Foundation (KF), and the publicly available Anatomical Tracings of Lesions After Stroke (ATLAS) dataset. To promote wide applicability, lesions were included from both subacute ($<$ 5 weeks) and chronic ($>$ 3 months) phases post stroke, and were of both hemorrhagic and ischemic etiology. Cross-study validation results (with independent training and validation datasets) were obtained to compare with previous methods based on naive Bayes, random forests, and three recently published convolutional neural networks. Model performance was quantified in terms of the Dice coefficient, a measure of spatial overlap between the model-identified lesion and the human expert-identified lesion, where 0 is no overlap and 1 is complete overlap. Training on the KF and MCW images and testing on the ATLAS images yielded a mean Dice coefficient of 0.54. This was reliably better than the next best previous model, UNet, at 0.47. Reversing the train and test datasets yields a mean Dice of 0.47 on KF and MCW images, whereas the next best UNet reaches 0.45. With all three datasets combined, the current system compared to previous methods also attained a reliably higher cross-validation accuracy. It also achieved high Dice values for many smaller lesions that existing methods have difficulty identifying. Overall, our system is a clear improvement over previous methods for automating stroke lesion segmentation, bringing us an important step closer to the inter-rater accuracy level of human experts.
\end{abstract}

\end{frontmatter}


\pagebreak

\section{Introduction}
\label{intro}
Neuropsychological studies of brain lesion-deficit relationships are an indispensable means of determining what brain areas are critical for carrying out particular functions. This contrasts with functional brain imaging techniques such as functional magnetic resonance imaging (fMRI), which while extremely popular and useful, cannot make strong claims about what brain areas are necessary for the functions being investigated. A major impediment to progress in brain lesion-deficit studies, however, is the labor-intensive and ultimately subjective step of having an expert manually segment brain lesions from MRI scans. 

This has been highlighted in previous studies comparing inter-rater variability and speed of human compared to automatic lesion identification. Fiez et al. \cite{fiez2000} report a 67\% ($\pm$ 7\%) agreement in overlapping voxels between two expert raters across ten subjects. 
More recently, other groups have reported an inter-rater overlap of 0.73 $\pm$ 0.2 between experts performing manual lesion segmentation for the ATLAS database \cite{liew2018large}. When brain lesion segmentation is performed exclusively by experienced neuroradiologists, median inter-rater agreement has been shown to be as high as 0.78 \cite{neumann2009interrater}. However, the involvement of only a small number of patients (N = 14) and the use of lower-resolution scans (6.5 mm slices rather than the typical 1 mm slices used in research) suggests that an inter-rater agreement of 0.78 may be inflated relative to the 0.67 to 0.73 range that seems typical for research studies.

Aside from concerns with inter-rater reliability, manually segmenting lesions is also time consuming, often taking between 4.8 to 9.6 hours. Methods developed for automating this process, however, can segment lesions in roughly a minute \cite{WILKE20112038}. However, manual lesion segmentation remains the method of choice, presumably due to the relatively poor accuracy of available automated methods \cite{WILKE20112038,ito2018comparison}. Clearly what is needed is a fast, automated method for brain lesion segmentation with a better accuracy than currently available methods.

Indeed, identifying lesions in brain MRI images is a key problem in medical imaging \cite{akkus2017deep,bernal2018deep}. Previous studies have examined the use of standard machine learning classifiers \cite{maier2015classifiers,griffis2016voxel,pustina2016automated} and convolutional neural networks (CNN) 
\cite{ronneberger2015u,he2016deep,guerrero2018white,kamnitsas2017efficient} for solving the problem of automating lesion segmentation. Machine learning methods like random forests tend to perform competitively \cite{maier2015classifiers} but fare below convolutional neural networks \cite{rachmadi2017deep}.

The first convolutional UNet \cite{ronneberger2015u} and subsequent models such as UResNet \cite{kamnitsas2017efficient} take as input 2D slices of the MRI image in a single orientation. They predict the lesion for each slice separately and then combine the predictions into a volume. This approach has limited accuracy because it does not consider the other two planes in the image volume. Without some method, such as a post-processing mechanism, for considering views from other orientations, models such as this will be inherently limited by how well a lesion can be detected in a single orientation view. For example, a wide and flat lesion might be readily distinguishable from healthy tissue in an axial but not coronal view. Indeed, a lesion that is more visible in sagittal and coronal views than in the axial view is shown in Figure~\ref{small_lesion}.

To address this limitation, CNN systems have been introduced that can accommodate multiple 2D slice orientations. The dual-path CNN, DeepMedic \cite{kamnitsas2017efficient}, while not considering multiple 2D orientations, does have two pathways, one for high and one for low resolution slices. Lyksborg et al. \cite{lyksborg2015ensemble} use a three path network, one for each of the canonical axial, sagittal and coronal views. Indeed, multi-path systems with up to eight different network paths have been explored previously \cite{de2015deep}. Adding paths, however, comes with a cost of having to fit many additional parameters for each path. Fitting these additional parameters leads to an increased risk of over-fitting, as has been reported for multi-path systems \cite{bernal2018deep}.

Multi-path systems must also combine the predictions from each path into a final output. One approach to combining path predictions is a simple majority vote. This was the approach used by Lyksborg et al. \cite{lyksborg2015ensemble}. However, this approach risks ignoring important but less frequently represented information, as the outputs from different paths are combined into a final voxel prediction by a simple majority vote. Also, the goal of their network was to segment tumors, where the pathology may present a somewhat different problem than stroke. Indeed, in the current work we show that majority vote performs less well on stroke lesion segmentation than a more inclusive 3D convolutional approach to combining outputs across paths.

We address shortfalls in previous approaches by proposing a novel nine-path system, where each path contains a custom U-Net to accommodate multiple MRI modalities or views, depending on the use case. For example, having both T1 and FLAIR modalities could be useful for segmenting sub-acute strokes that have occurred within, say, the last 5 weeks. For more chronic strokes having occurred more than 6 months previous, multiple T1 views might be more useful than combining with FLAIR. This possibility is tested in Table 1 below. Our system considers three different normalizations of the images along each of the three axial, sagittal and coronal views. Our custom U-Net is weak on its own but powerful as a component of our multi-path system. This makes sense in the context of ensemble learning where weak learners can perform better in an ensemble \cite{freund1997decision}. We also use a 3D convolutional kernel to merge 2D outputs from each path and show that it gives a better accuracy than majority vote. It is because of this combination of 2D and 3D approaches that we refer to our system as 2.5D.

Critically, we address the challenging issue of model over-fitting by performing a rigorous cross-study validation to evaluate accuracy of lesion identification across sites that differ in numerous ways such as scanner model, patient sample, and expert tracers. This is done by training a model on one set of patient MRIs and then testing the ability of those trained parameters to identify lesions in a separate validation (test) set. Cross-study validation gives a better estimate of the model's true accuracy compared to cross-validation, where train and test samples are simply re-shuffled from the same dataset \cite{bernau2014cross}. 

Details of our model are provided below, followed by experimental results across three different datasets. We show that our system has significantly higher agreement with ground-truth segmentations by human experts compared to the recent CNN-based methods DeepMedic \cite{kamnitsas2017efficient}, the original UNet \cite{ronneberger2015u}, a residual UNet \cite{guerrero2018white}, and two non-CNN based machine learning methods based on random forests \cite{pustina2016automated} and naive Bayes \cite{griffis2016voxel}.

\section{Methods}
\label{methods}

 
\subsection{Convolutional neural networks}
Convolutional neural networks are the current state of the art in machine learning for image recognition \cite{lecun1998gradient,krizhevsky2012imagenet}, including for MRI \cite{bernal2018deep}. They are typically composed of 
alternating layers for convolution and pooling, followed by a final flattened layer. A convolution layer is specified by a filter size and the number of filters in the layer. Briefly, the convolution layer performs a moving dot product against pixels given by a 
fixed filter of size $k \times k$ (usually $3\times3$ or $5\times5$). The dot product is 
made non-linear by passing the output to an activation function such as a sigmoid or rectified linear unit (also called relu or hinge) function. Both are differentiable and thus fit into the standard gradient descent framework for optimizing neural networks during training. The output of applying a $k\times k$ convolution against a $p\times p$ image is an image of size $(p-k+1) \times (p-k+1)$. In a CNN, the convolution layers just described are typically alternated with pooling layers.
The pooling layers serve to reduce dimensionality, making it easier
to train the network. 

\subsection{Convolutional U-network}
After applying a series of convolutional filters, the final layer dimension is usually much smaller than that of the input images. For the current problem of determining whether a given pixel in the input image is part of a lesion, the output must be of the same dimension as the input. This dimensionality problem was initially solved by taking each pixel in the input image and a localized region around it as input to a convolutional neural network instead of the entire image \cite{ciresan2012deep}. 

A more powerful recent solution is the Convolutional U-Net (U-Net) \cite{ronneberger2015u}. This has two main features that separate it from traditional CNNs: (a) deconvolution (upsampling) layers to increase image dimensionality, and (b) connections between convolution and deconvolution layers. Another popular U-Net method is the residual U-Net (also known as UResNet \cite{guerrero2018white}) that has residual connections to prevent the gradient from becoming zero (also called the vanishing gradient problem \cite{hochreiter1998vanishing}). 


\subsection{U-Net systems}
Since the introduction of the original U-net, several systems have been proposed for analyzing MRI images. DeepMedic is a popular multi-path 3D CNN model that combines high and low resolutions of input images. Previous systems like Lyksborg et. al. \cite{lyksborg2015ensemble} consider the three axial, sagittal, and coronal planes in a multi-path ensemble, but use a potentially limiting majority vote approach to combine outputs from each path. Multi-path systems can be challenging to train, as can be seen in the work of Brebisson and Montana \cite{de2015deep}. There they train eight networks in parallel to capture various aspects of the input image but report overfitting due to large number of parameters. 

Post processing is another important component of U-Net systems to reduce false positives. The post processing methods range from simple ones like connected components and clustering \cite{lai2015deep,havaei2017brain} to using 3D CNNs and conditional random fields \cite{kamnitsas2017efficient}. The latter methods also end up accounting for temporal dependence between slices, resulting in a higher accuracy. 


\subsection{Our CNN system}
\subsubsection{Overview}
\begin{figure}[h]
  \centering
  \begin{tabular}{cc}
\includegraphics[scale=.13]{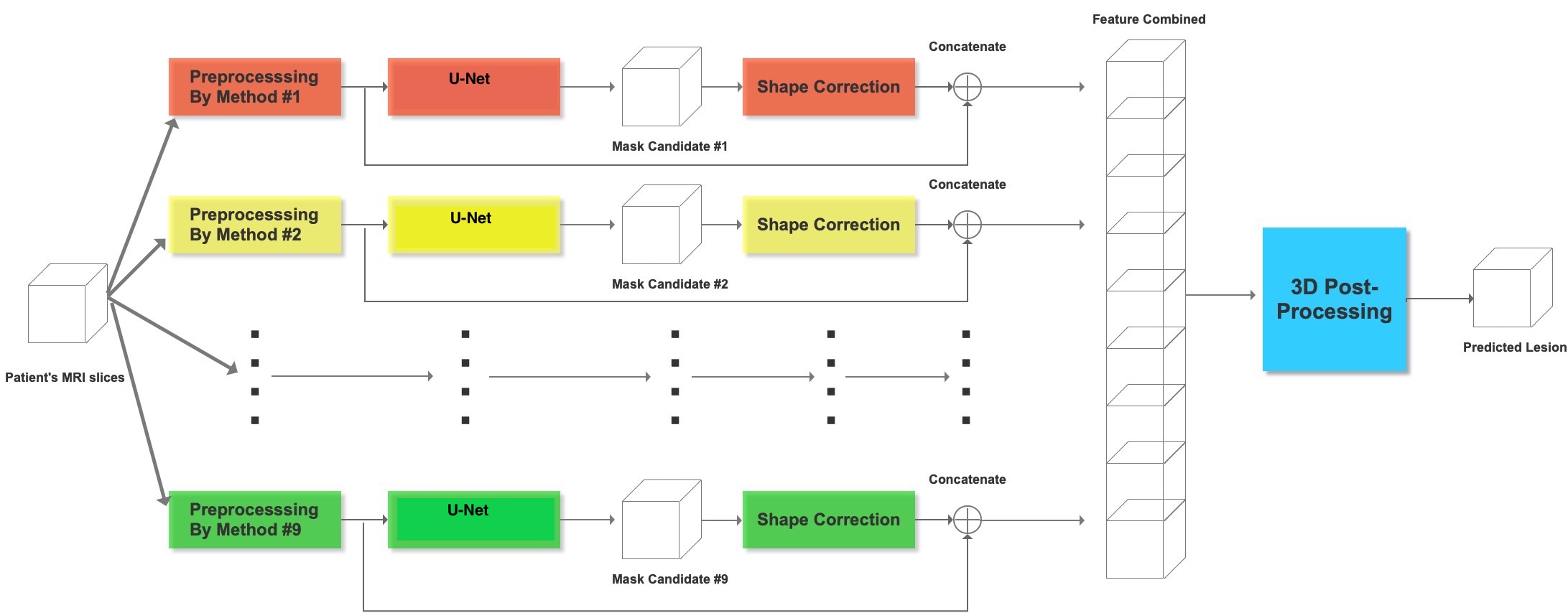} &  \includegraphics[scale=.17]{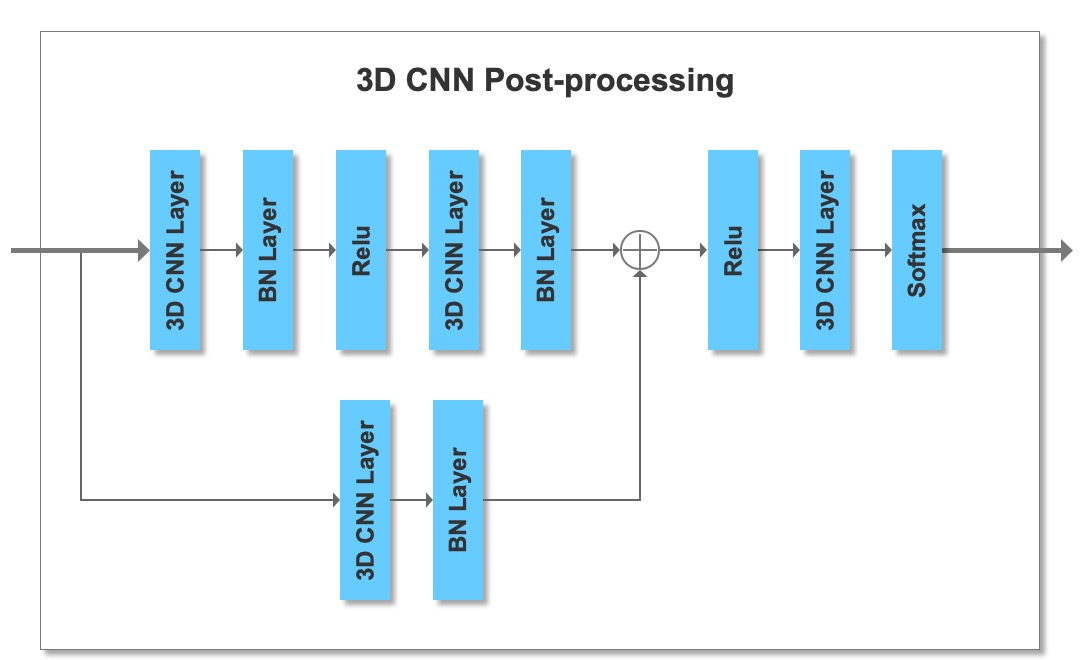}  \\
\footnotesize A  & B \footnotesize  \\
\end{tabular}
  \caption{Overview of our entire nine-path system (A) and a zoomed in view of our 3D CNN post processor (B) for combining outputs from each path. \label{overview} }
\end{figure}

We developed a modified U-network in a multi-path multi-modal system with a 3D convolutional kernel for post-processing shown in Figure~\ref{overview}. A 3D kernel is like a 2D one except that it has a third dimension that it convolves into as well, and thus it expects a 3D input. For example, in a 2D system kernels are typically $3\times3$ whereas in a 3D kernel it would be $3\times3\times3$. Details of our system are provided below, highlighting differences in our approach compared to previous ones.

\subsubsection{Multiple paths}
Our primary motivation for taking a multi-path approach is to optimize the ability of the model to identify brain lesions by capturing image information from all three angles as well as their normalizations. To return to the overview of our system shown in Figure~\ref{overview}(a), consider the three different normalizations of each of the three axial, sagittal, and coronal planes. For each plane we normalize (1) in the same plane, (2) across the third plane, and (3) both in the same plane first and then across the third, thus giving nine paths. These choices were motivated by our preliminary results not shown here and previous studies showing that different planes work best for different lesion locations \cite{lyksborg2015ensemble}, and that the best method of normalization may differ depending on image view \cite{bernal2018deep}.

\subsubsection{Basic U-net}
\begin{figure}[!h]
  \centering
\includegraphics[scale=.2]{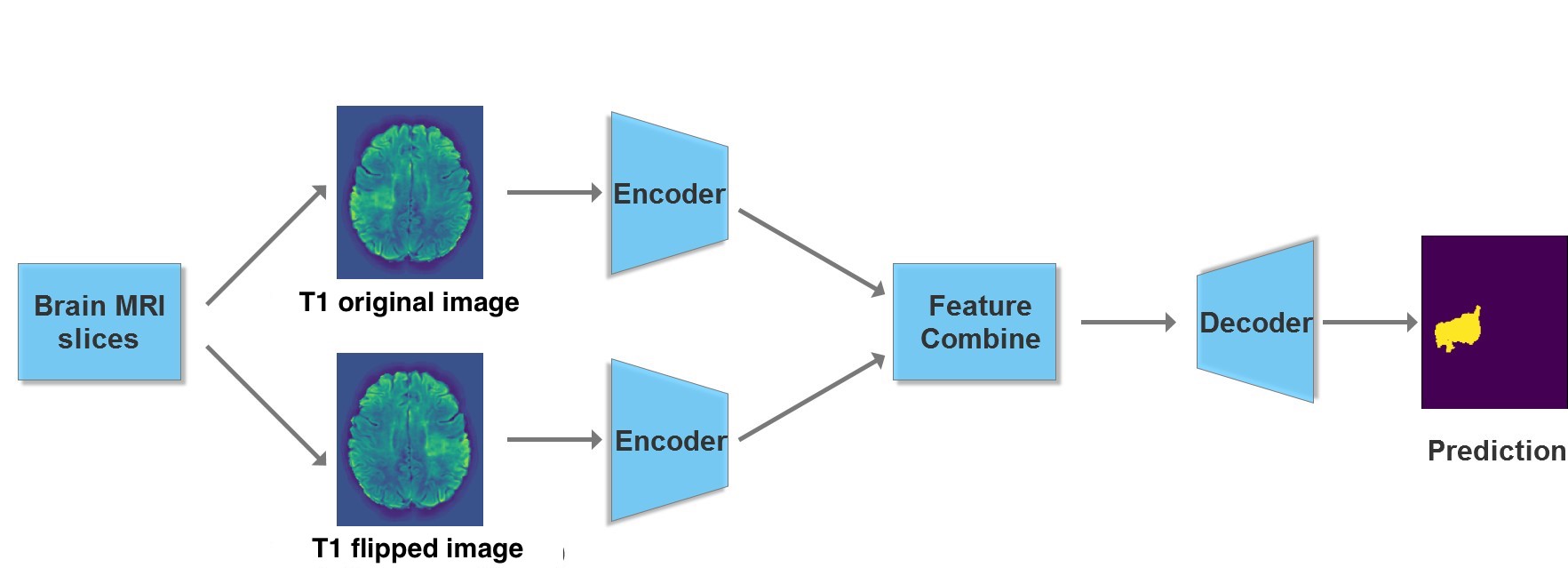}  \\
\footnotesize (a) Overview of our dual-path U-network. We have a separate encoder for the original T1 image of the brain scan and one for its flipped version. Alternatively, two different image modalities may also be used instead of two different hemispheres.  \\
 \begin{tabular}{cc}
 \includegraphics[scale=.14]{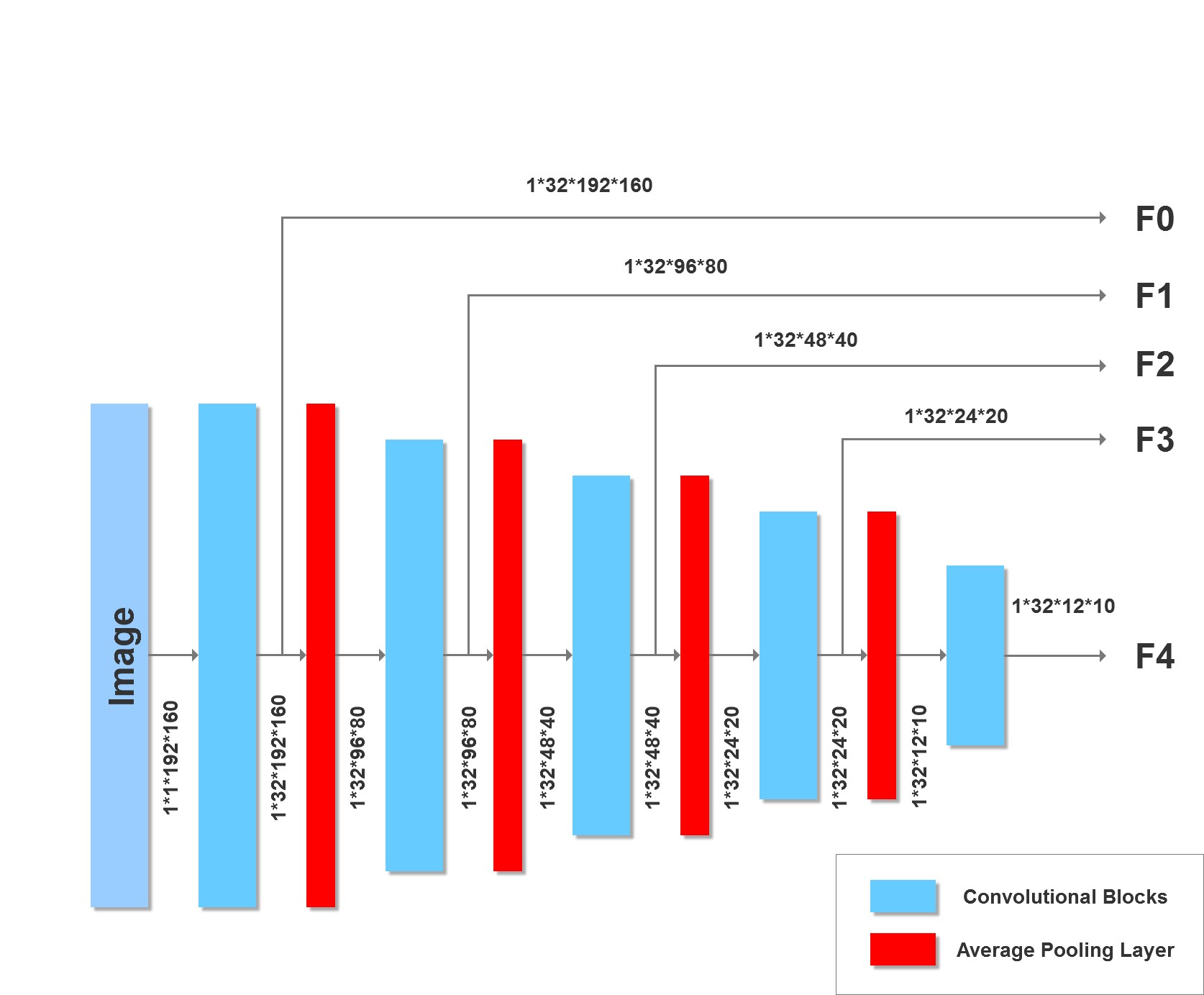} & \includegraphics[scale=.14]{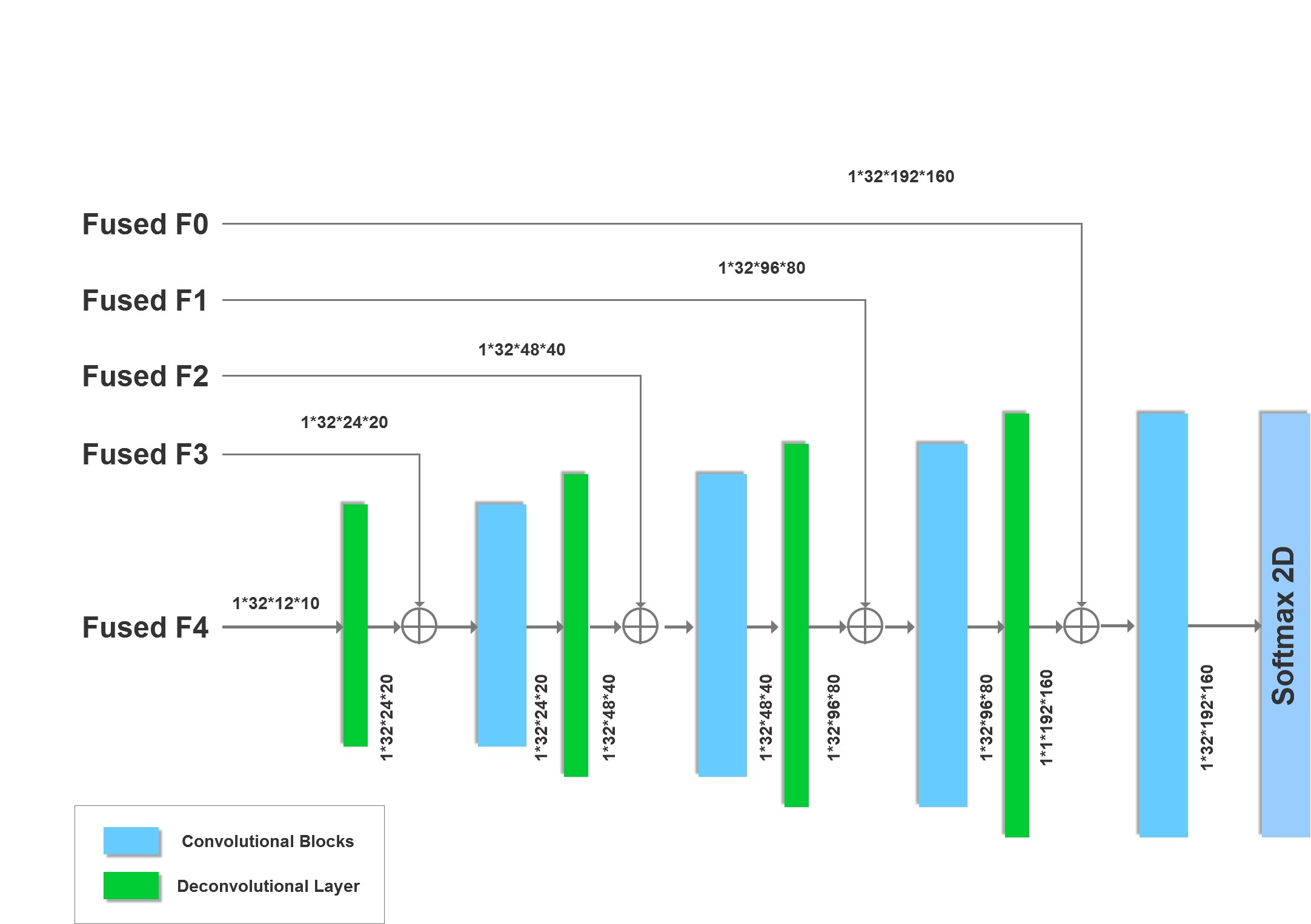} \\
\footnotesize (b) U-Net Encoder with five convolutional blocks 
& \footnotesize (c) U-Net Decoder with four convolutional and deconvolutional blocks \\
\footnotesize Also shown are image dimensions after each convolution. &  
\footnotesize Also shown are image dimensions after each deconvolution. \\
\includegraphics[scale=.2]{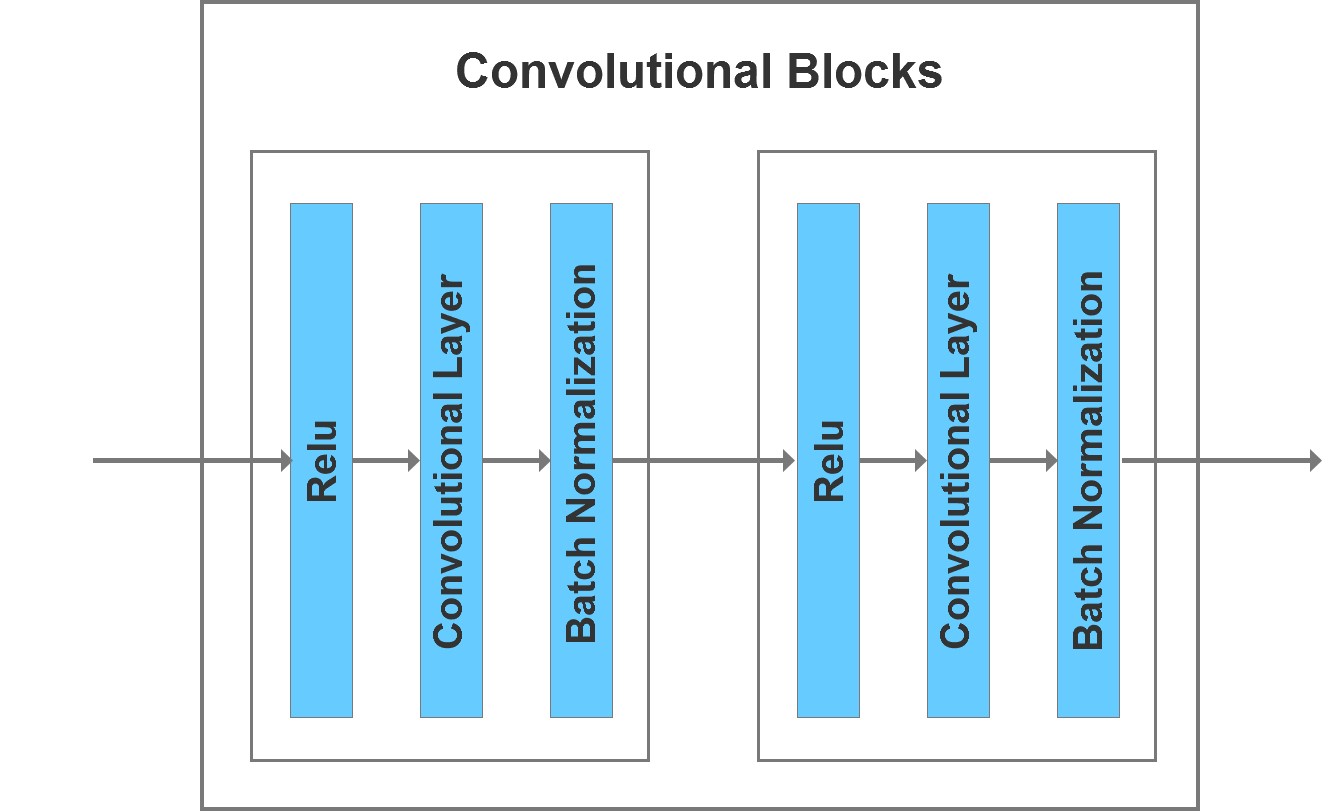}  & \includegraphics[scale=.1]{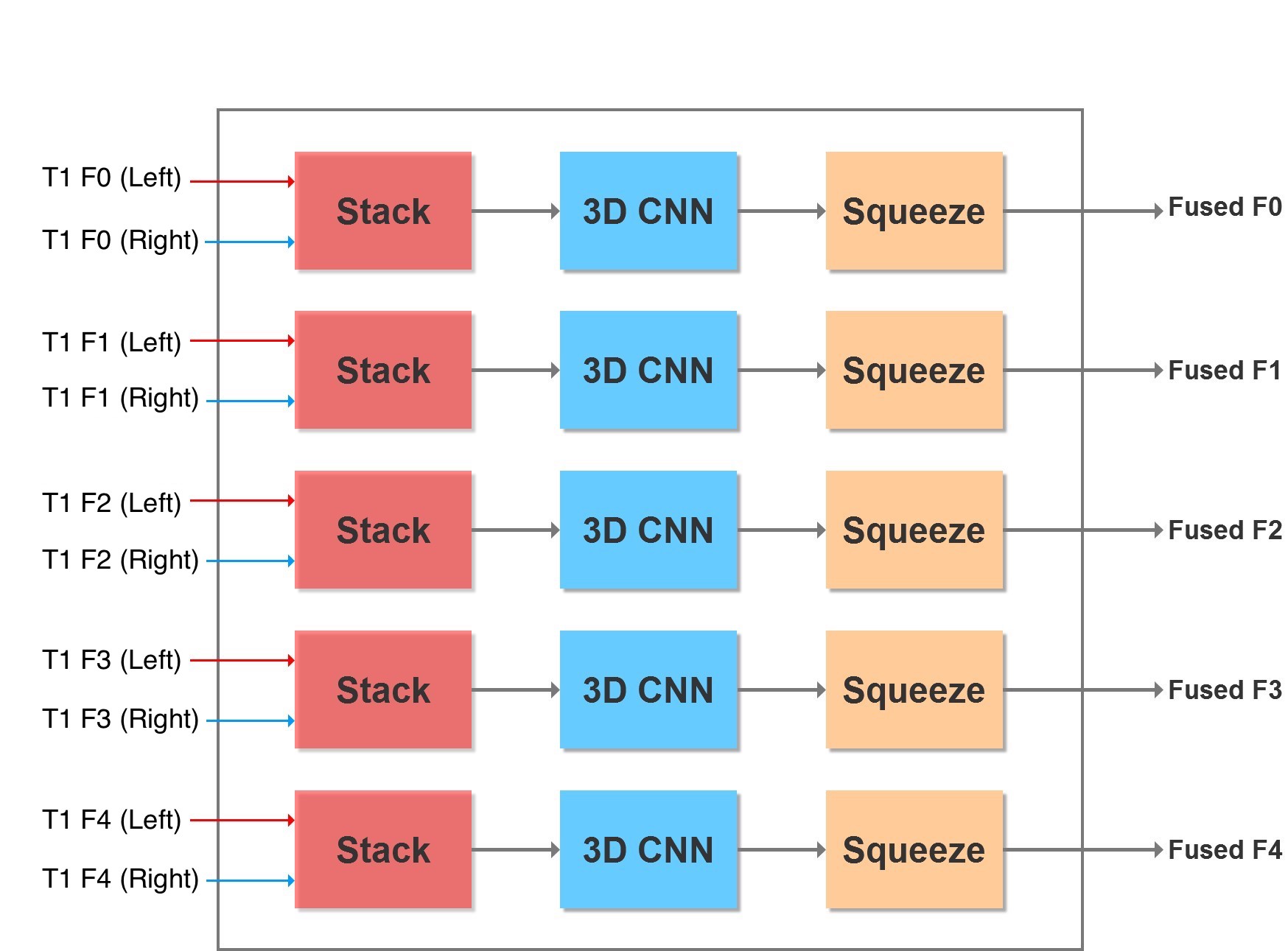} \\
\footnotesize (d)  Convolutional blocks used in encoder above & \footnotesize (e) Fuse features from encoding the original and flipped images \\
& \footnotesize (or alternatively encoding from two different image formats) \\
\end{tabular}
\caption{Our U-network  models with encoder and decoder details. \label{path}}
\end{figure}

\paragraph{Encoder} First we look at details of our basic U-net that makes up the system. Our U-net used in 
each path is inspired by the original U-net \cite{ronneberger2015u} and a more recent one 
\cite{tseng2017joint} that attains state of the art accuracies on the BRATS brain tumor MRI benchmark \cite{menze:hal-00935640}. The encoder portion of our U-net is shown in Figure~\ref{path}(c). After each convolution we perform a $2\times2$ average pooling with stride 2 to halve the image dimension. Features from the encoder are passed to the decoder. However, since there are two encoders (one for the original T1-weighted image and the other for its flipped version), corresponding features are combined using the block shown in Figure~\ref{path}(e). Alternatively, the current network can be used with two different MRI modalities by substituting the T1 image and its flipped version with separate left hemisphere T1-weighted and Fluid-Attenuated Inversion Recovery (FLAIR) images. 

\paragraph{Feature fusion} From each encoder we obtain a prediction of a lesion (in the respective normalization and plane) that we merge with a $2\times1\times1$ 3D convolutional kernel \cite{tseng2017joint,lai2015deep}. We take the two feature maps each of dimension  $32\times x \times y$ where 32 is the number of convolutional filters from the encoder layer and $x\times y$ is the input size depending upon the encoder layer (see Figure~\ref{path}(a)). Stacking refers to adding an extra dimension to make the input $32 \times 2 \times x \times y$ for the 3D kernel. The $2\times1\times1$ 3D kernel gives an output of  $32 \times 1 \times x \times y$ which is "squeezed" to remove the unnecessary dimension to give an output of $32 \times x \times y$ to the decoder.

\paragraph{Decoder} The fused features are then given to the decoder, which we add to the output of deconvolutional layers (briefly explained below), a process shown as a $\oplus$ sign in Figure~\ref{path}(c). The image dimensions are preserved because of the addition. The previous U-net that served as a starting point for our current effort  \cite{tseng2017joint} performed element-wise multiplication of fused features with deconvolved ones. However, this is unlikely to be useful for the current system. Our fused features and upsampled features have small values, so their product would even be smaller. This in turn would give a gradient with zero or near-zero values that would affect the training. Thus we prevent this by adding instead of multiplying fused and upsampled feature values. 

\paragraph{Convolutional blocks} Shown in blue in Figure~\ref{path}(d) are the convolutional blocks used in our encoder and decoder. We use $3\times3$ convolutional blocks with a stride of 1 and padding of one extra layer in the input to make the output dimensions same as the input. The previous U-net that inspired our design \cite{tseng2017joint} performed Relu activation before adding fused features. Here we perform Relu activation twice. In the context of the decoder, this means Relu activation is performed after adding fused features to upsampled ones. Performing Relu activation after addition rather than before has been shown to be more accurate for image classification \cite{he2016identity}.

\paragraph{Deconvolutional blocks} Deconvolutional blocks (also known as transposed or fractionally strided convolutions) are meant to increase the dimensionality of images \cite{dumoulin2016guide}. The term transpose arises from the fact that a deconvolution is simply the product of the transpose of the convolution weight matrix with the output when the stride is 1. If the stride is more than one we insert zeros in between the input to obtain the correct transpose result (as well-explained in Dumoulin and Visin \cite{dumoulin2016guide}) We use $2\times2$ deconvolutions with a stride of 2 that doubles the image dimensions in both axes. 

\subsubsection{Post-processing}
The output of each of the nine paths in our system is a 2D mask showing the predicted location of the lesion in the same view as the input image, as in Figure~\ref{path}(a). The lesion prediction mask is binarized by rounding to 0 if the values in the mask are below 0.5, otherwise values are rounded up to 1. We stack each predicted lesion with the original input image and combine all slices to form a $2\times192\times224\times192$ volume. Since we have nine paths this becomes of size $18\times192\times224\times192$. This is passed to our 3D CNN post-processor as described below.

In the post-processor shown in Figure~\ref{overview}(b), we have a main path containing 36 3D $3\times3\times3$ kernels each with 18 channels, or equivalently 36 3D kernels each of size $18\times3\times3\times3$. Following that, the second 3D CNN in the main path has 9 3D $3\times3\times3$ kernels each with 36 channels, and two final 3D CNNs each of dimensions $3\times3\times3$ with 9 channels. 

\subsubsection{Loss function}
The final output from the post-processor has two channels each of dimensions $192\times224\times192$. The target lesion has the same dimensions but just one channel. The first channel in our output predicts the lesion and the second one predicts the complement of it. We convert the outputs of each channel into probabilities with softmax \cite{alpaydin} and combined them into a modified Dice loss function \cite{milletari2016v,wong20183d}. For a single channel output the Dice loss is defined to be $1-D$ where

\begin{center}
$D(p) = \frac{2\sum_i p_i r_i}{\sum_i p_i^2 + \sum_i r_i^2}$
\end{center}

$p_i$ are the predicted softmax outputs of the channel, and $r_i$ is 1 if the voxel has a lesion and 0 otherwise. If we are predicting the complement of the lesion then the values of $r_i$ are flipped from 0 to 1 and 1 to 0. With our two channel output $p$ and $q$ our loss becomes $2-(D(p)+D(q))$ where the latter $D(q)$ is for the complement.

\subsection{Imaging Data}
We obtained high-resolution (1 mm$^{3}$) whole-brain MRI scans from 25 patients from the Kessler Foundation (KES), a neuro-rehabilitation facility in West Orange, New Jersey. We also obtained 20 high-resolution scans from the Medical College of Wisconsin (MCW). Data heterogeneity is important for widespread applicability of the model. To that end, we included data from a variety of time points: subacute ($<$ 5 weeks post stroke) and chronic ($>$ 3 months post stroke). Strokes of both hemorrhagic and ischemic etiology were included. The lesions visualized on the scans were hand-segmented by a trained human expert, as described for the KES scans in \cite{boukrina2015neurally} and the MCW scans \cite{pillay2014cerebral,binder2016surface}. To move these scans into standard Montreal Neurological Institute (MNI) reference space \cite{fonov2011unbiased}, we used the non-linear warping tool, 3dQwarp, from the AFNI software suite \cite{cox1996afni}. The segmented lesion was used as an exclusion mask so that the lesioned territory would be excluded from the warping procedure. This prevents non-lesioned brain tissue from being distorted to fill in the lesioned area. This transformation was performed on the T1 images, resulting in skull-stripped output in MNI space. This calculated transformation for each participant was then applied to the FLAIR image (KES only) and hand-traced lesion mask.

We also obtained scans and stroke lesion masks from the public ATLAS database \cite{liew2018large} and processed them as just described for the KES and MCW data. We selected images according to the following criteria to focus on cases with single lesions in the left hemisphere:

\begin{verbatim}
Session = t01 [T1 Scans only]
LH_Cort + LH_SubCort = 1 [Cortical OR Sub-Cortical Lesion only]
RH_Cort = 0 [No Cortical Right Hemisphere Lesion]
RH_SubCort = 0 [No Sub-Cortical Right Hemisphere Lesion]
Other_Location = 0 [No Lesion elsewhere]
Hemisphere = Left [Left Hemisphere only]
\end{verbatim}

This resulted in 54 images being selected from the ATLAS set. Thus we included a total of 99 images altogether across the three datasets. We divided these into two groups, ATLAS or Kessler+MCW, for cross-study comparisons. We then combined them to perform a five-fold cross-validation across all 99 images.

\subsection{Comparison of CNN Methods}
We compared our CNN to three state of the art recently published CNNs shown below. Our system was implemented using Pytorch \cite{paszke2017automatic}, the source code for which is available on our GitHub site \url{available.upon.acceptance}. In each of our experiments we train our model, UNet, and UResNet with stochastic gradient descent and Nesterov momentum \cite{ruder2016overview} of 0.9 and weight decay of .0001. We use a batch size of 32, starting from an initial learning rate of 0.01 with a 3\% weight decay after each epoch for a total of 50 epochs. In DeepMedic we use the default settings of learning rate of 0.001, the RMSProp optimizer \cite{ruder2016overview} with a weight decay of .0001, batch size of 10, and a total of 20 epochs.

\begin{itemize}
\item DeepMedic \cite{kamnitsas2017efficient}: This is a popular dual-path 3D convolutional neural network with a conditional random field to account for temporal order of slices. DeepMedic contains a path for low- and a separate path for high-resolution of images. Its success was demonstrated by winning the ISLES 2015 competition to identify brain injuries, tumors, and stroke lesions. The code for implementing DeepMedic is freely available on GitHub, \url{https://github.com/Kamnitsask/deepmedic}.
\item UResNet \cite{guerrero2018white}: This is a convolutional neural network with residual connections \cite{he2016deep}. The code for implementing UResNet is also freely available on GitHub, \url{https://github.com/DeepLearnPhysics/pytorch-uresnet}. 
\item UNet \cite{ronneberger2015u}: The was the original convolutional U-network proposed for biomedical image processing. Its code is also available on GitHub, \\
\url{https://github.com/thonycc/PFE/tree/af9e804f71684b73cf7f3b25557edcf6a1b307b3}.
\end{itemize}

Two other non-CNN-based machine learning packages were also included because they have been made freely available to the brain imaging community and have been developed for ease of use. Both take a patch-based approach to automating lesion segmentation. That is, these methods convert the input image into multiple patches that are used to train the model. They are LINDA \cite{pustina2016automated}, based on a random forests algorithm, and a second method based on Gaussian naive Bayes \cite{griffis2016voxel}.

\subsection{Data analysis}
\subsubsection{Measure of accuracy: Dice coefficient}
The Dice coefficient is typically used to measure the accuracy of predicted lesions in MRI images \cite{zijdenbos1994morphometric}. The output of our system and that of other methods is a binary mask of the dimensions as the input
image, but with a 1 for each voxel calculated to contain a lesion, and a 0 otherwise.
Comparison of the human expert-segmented lesion mask with that from automated methods is quantified with the Dice coefficient. Starting with the human binary mask as ground truth, each predicted voxel is determined to be either a true positive (TP, also one in true mask), false positive (FP, predicted as one but zero in the true mask), or false negative (FN, predicted as zero but one in the true mask). The Dice coefficient is formally defined as

\begin{equation}
DICE=\frac{2TP}{2TP+FP+FN}
\end{equation}

\subsubsection{Measure of statistical significance: Wilcoxon rank sum test}
The Wilcoxon rank sum test \cite{wilcoxon1945individual} (also known as the Mann-Whitney U test) can be used to determine whether the difference between two sets of measurements is significant. More formally, it tests for the null hypothesis that a randomly selected point from a sample is equally likely to be lower or higher than a randomly selected one from a second sample. It is a non-parametric test for whether two sets of observations are likely to be from different distributions, without assuming a particular shape for those distributions.

\section{Results}
In the results presented below, we take the rare and rigorous step of performing cross-study validations across independent datasets \cite{bernau2014cross}. We also examine results from cross-validation in the combined dataset from the three different sources (KES, MCW, and ATLAS).

\subsection{Cross-study validation results}
To create relatively balanced sets in terms of number of scans, we combine the KES and MCW datasets into one. This yielded 45 samples in KES+MCW and 54 in ATLAS. We first train all convolutional neural networks (CNNs) on the KES+MCW data and test their ability to predict lesion locations in the ATLAS set. We then repeat the same procedure but with the train and test datasets reversed. Since LINDA and GNB come pre-trained and were intended for out-of-the-box use rather than re-training, we ran them as-is. Both programs have skull-removal built into their pipelines. Because the ATLAS images were the largest dataset with the skull still intact, we restricted our test of the LINDA and GNB methods to the ATLAS dataset.

\subsubsection{Train on KES+MCW, predict on ATLAS}
\begin{figure}[h]
\centering
\includegraphics[scale=.3]{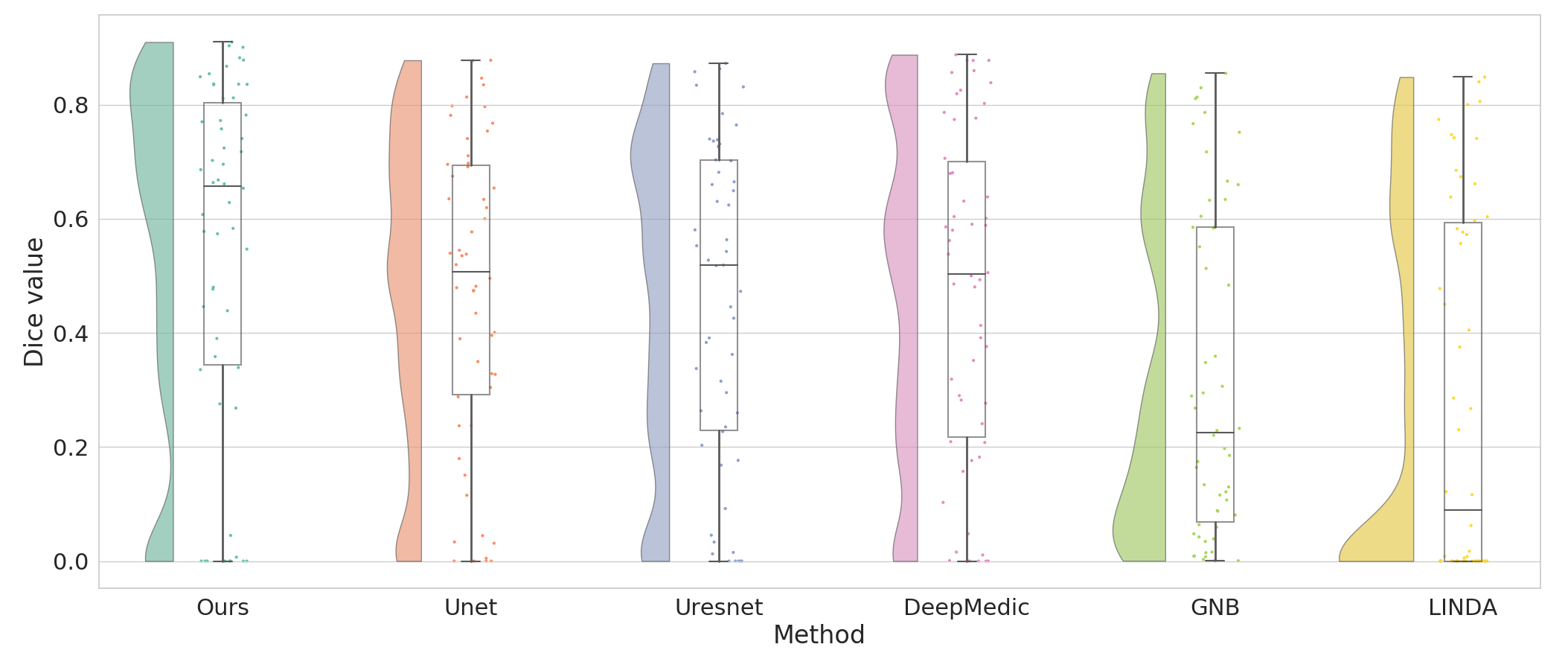} \\
\centering
\footnotesize
\begin{tabular}{ccccccccc}\hline
Method & Our system & UNet & UResNet  & DeepMedic & LINDA & GNB \\ 
Mean Dice & 0.54 & 0.47 & 0.45 & 0.47 & 0.32 & 0.29 \\  \hline
\end{tabular}
\caption{Raincloud plots of Dice coefficient values of all models trained on KES+MCW and tested on ATLAS. For each method we show the distribution of Dice coefficients across all test images as well as the five summary values: median (middle horizontal line), third quartile (upper horizontal line), first quartile (lower horizontal line), min (lowermost bar), and max (uppermost bar). All models except for LINDA and GNB are trained on KES+MCW. The Table below the graph contains the mean Dice coefficients of all models on the ATLAS test data. 
\label{atlas}}
\end{figure}

Figure~\ref{atlas} shows the Dice coefficient values on the ATLAS test dataset with training performed on KES and MCW images. Results show that the current system, with a median Dice value of 0.66, yielded the best performance. This was not just due to a few high values, as its Dice values generally clustered toward the higher end. 
The Dice values of UNet, UResNet, and DeepMedic Dice have a more even distribution than our system and
lower median values. Both LINDA and GNB have Dice values clustered toward the lower end. Figure~\ref{atlas} also shows that our system has the highest mean Dice value. This value is reliably higher than all other methods under the Wilcoxon rank test \cite{wilcoxon1945individual} (p $<$ 0.001). All the convolutional networks achieve better median values than LINDA and GNB. 

\subsubsection{Train on ATLAS, predict on KES+MCW}
\begin{figure}[h!]
\centering
\includegraphics[scale=.3]{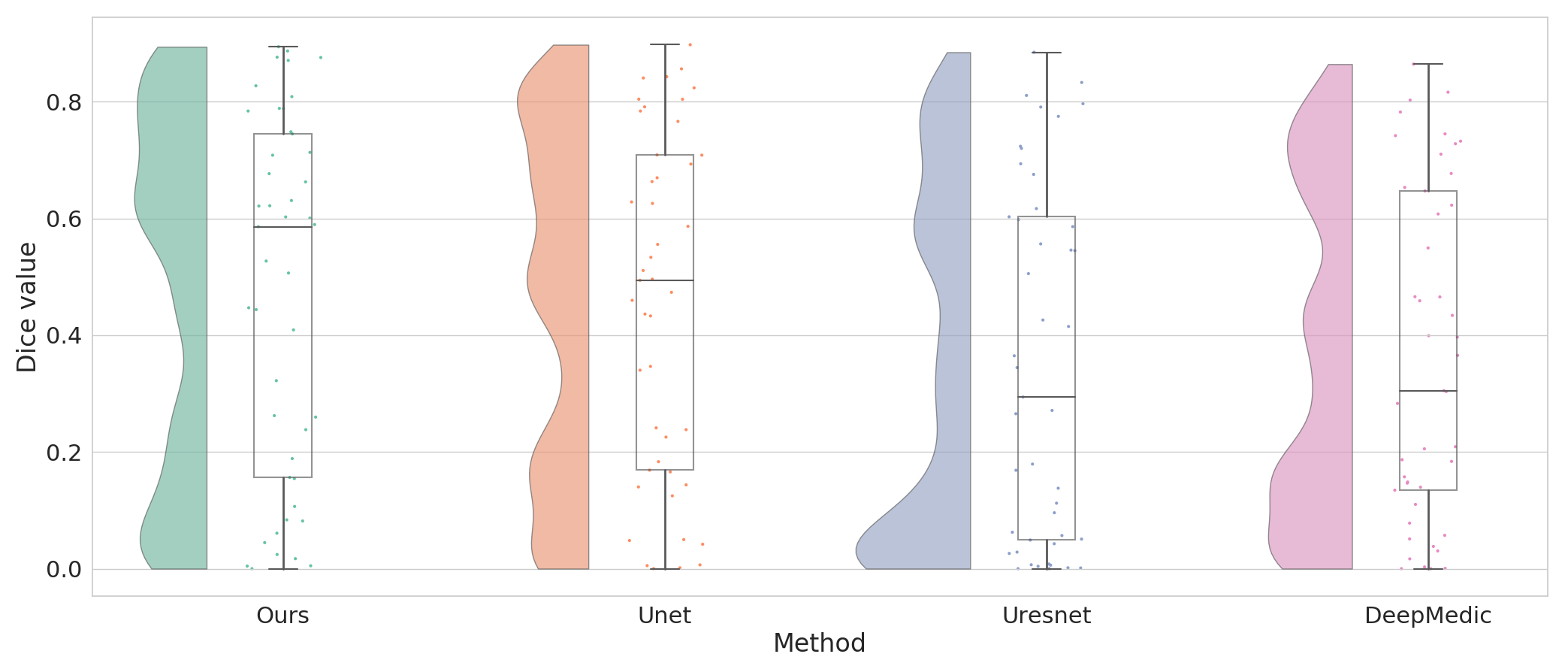} \\
\footnotesize
\begin{tabular}{cccccccc}\hline
Method & Our system & UNet & UResNet  & DeepMedic \\ 
Mean Dice & 0.47 & 0.45 & 0.35 & 0.37 \\  \hline
\end{tabular}
\caption{Raincloud plots of Dice coefficient values for all models trained on ATLAS and tested on KES+MCW. Also shown in the table are mean Dice coefficients of each method, as tested on the KES+MCW set.
\label{kcw}}
\end{figure}

Figure~\ref{kcw} shows results from the other direction of the cross-study analysis: training on ATLAS and testing on KES+MCW. In this case, although our system has the highest median, its distribution of Dice values is no longer clustered toward the high end as it was previously. The mean Dice value of our system is marginally
above that of UNet alone and not statistically distinguishable from it. Compared to UResNet and DeepMedic, however, our method performs better, as shown from its reliably higher Dice values (p $<$ 0.001).

\begin{figure}[h!]
\centering
\includegraphics[scale=.3]{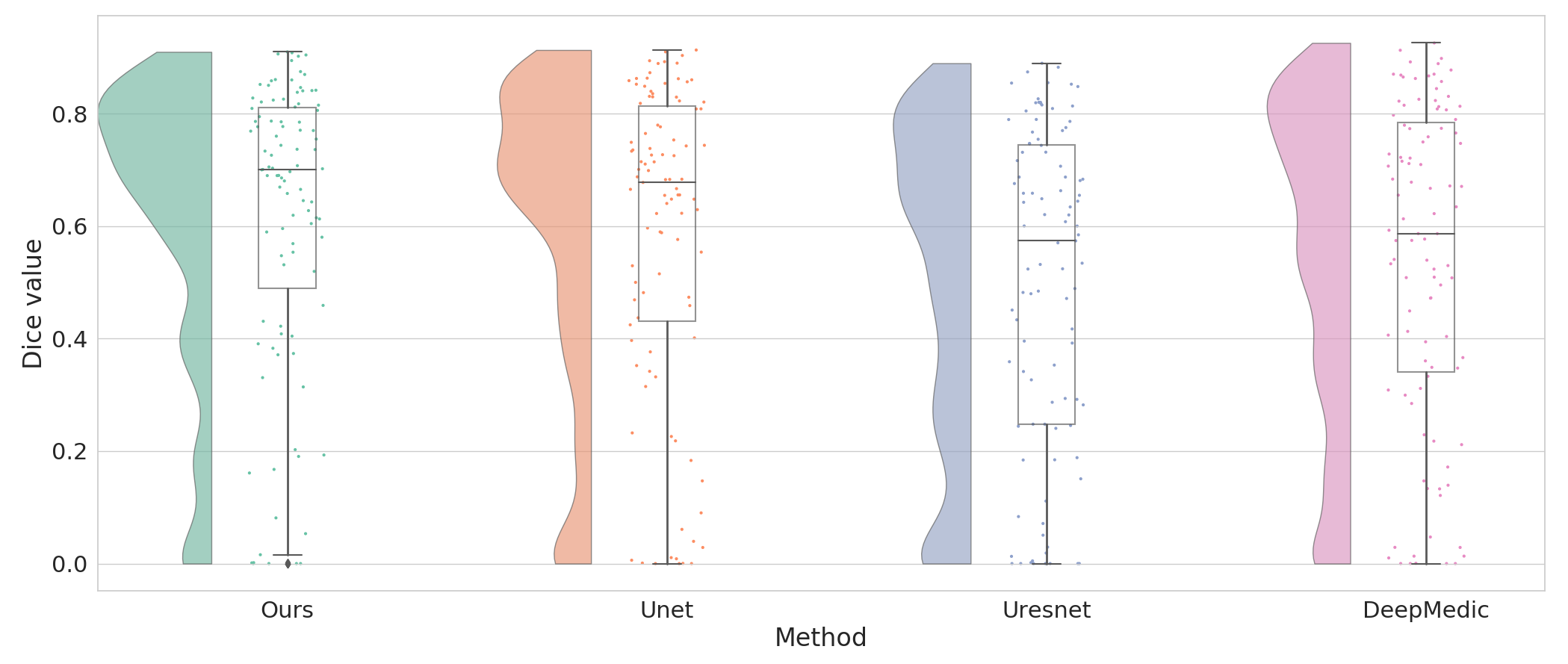} \\
\centering
\footnotesize
\begin{tabular}{cccccccc}\hline
Method & Our system &  UNet & UResNet  & DeepMedic \\ 
Mean Dice & 0.62 & 0.58 & 0.49 & 0.54 \\  \hline
\end{tabular}
\caption{Raincloud plot of Dice coefficient values obtained by five-fold cross validation on all our data combined: ATLAS+Kessler+MCW. In the Table are the mean Dice coefficients given by cross-validation. \label{cv}}
\end{figure}

\subsection{Cross-validation results on all datasets ATLAS, KES, and MCW combined}
To take full advantage or our relatively large dataset, we combined images from all three sources to produce an overall dataset of 99 samples. We then performed a five-fold cross-validation on this combined dataset to evaluate the accuracy of each method. Figure~\ref{cv} shows that our system again has the highest median Dice value. Our system also has the highest mean Dice value at 0.62, performing reliably better than the next best system, UNet, at 0.58. Indeed, our system performed better (p $<$ 0.001) than all three of the other CNN-based systems.

\begin{wrapfigure}{r}{0pt}
\vspace{-20pt}
  \centering
    \includegraphics[scale=.78]{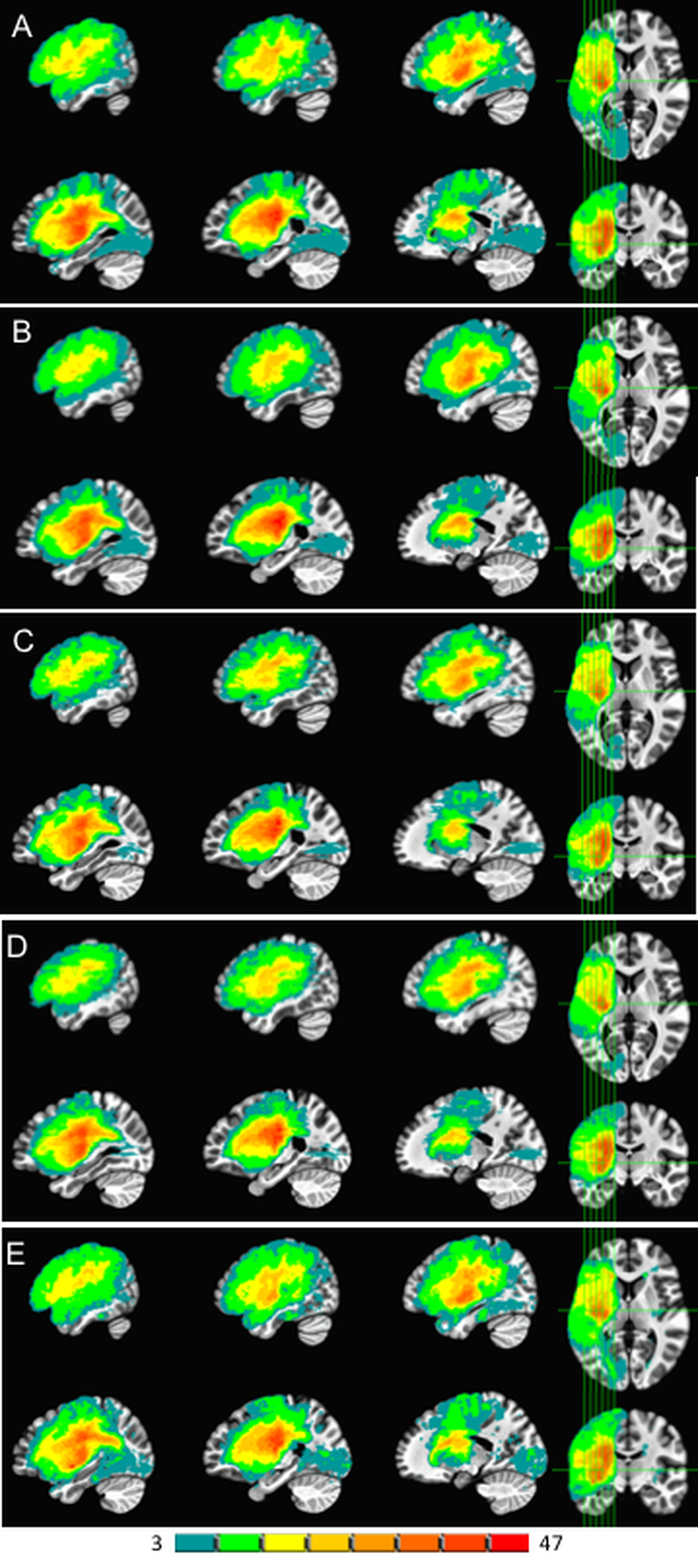}
    \vspace{-20pt}
\caption{Lesion overlap map results from 5-fold cross-validation on the entire 99 scan dataset. The leftmost side of the color scale in teal shows locations with 3 spatially overlapping lesions, while the rightmost side in red shows a maximum of 47 overlapping lesions. Hand-segmented lesions are in panel A. Our 2.5D CNN model is in panel B. The UNet model is in panel C. The URestNet model is in panel D. And the DeepMedic model output is in E. \label{overlaps}}
\vspace{-25pt}
\end{wrapfigure}

In addition to reporting this advantageous numeric performance of our system, an overall illustration of how the lesion masks produced by the current model compared to those from the other CNN-based models is in Figure~\ref{overlaps}. The expert-traced lesions (A) are shown alongside those produced by our system (B) and the models (C-E). 
One point to note is that while our system performed significantly better in terms of overlap with human expert tracings as measured by the Dice coefficient, visually all the automatic methods appear grossly similar to the human expert segmentations.

\subsection{Distribution of Dice coefficients across lesion size}
\begin{figure}[h]
\centering
\includegraphics[scale=.3]{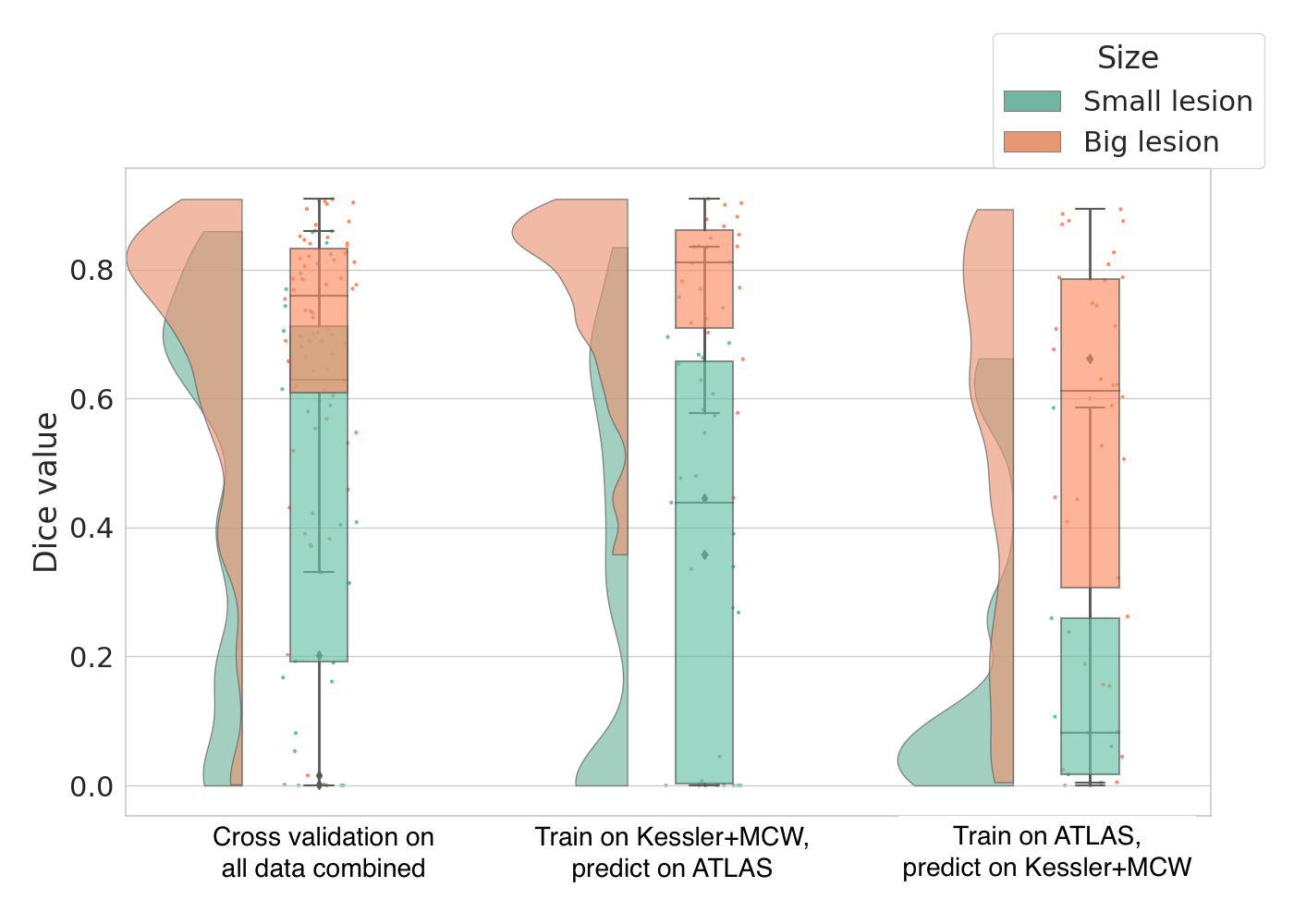}
\footnotesize
\begin{tabular}{cccccccc}\hline
Method & Our system & UNet & UResNet  & DeepMedic \\ 
Small lesions & 0.5 & 0.48 & 0.36 & 0.33 \\  
Large lesions & 0.69 & 0.66 & 0.59 & 0.68 \\ \hline
\end{tabular}
\caption{Raincloud plot showing the distribution and five summary statistics of Dice coefficients in three different scenarios. The left panel shows Dice values given by cross-validation on all the data combined. The middle panel shows a cross-study scenario where the current model is trained on KES+MCW and tested on ATLAS. The right panel shows results from training on ATLAS and testing on KES+MCW. In the Table below the plots we show the mean Dice values of our system and the other CNNs on small and large lesions separately. \label{lesionsize}}
\end{figure}

Lesions with $x\times y\times z$ dimensions less than $20\times20\times25$ mm were classified as small, and any lesions with dimensions greater than those were considered large. In Figure~\ref{lesionsize} we show a raincloud plot of Dice values obtained by our system in the cross-validation and cross-study settings.

Smaller lesions are generally harder to identify than larger ones \cite{griffis2016voxel,pustina2016automated,ito2018comparison}. To compare performance between lesion sizes, we split the lesions into small and large categories based on the distribution of lesion sizes in the overall set. 

\begin{figure}[h]
\includegraphics[scale=.93]{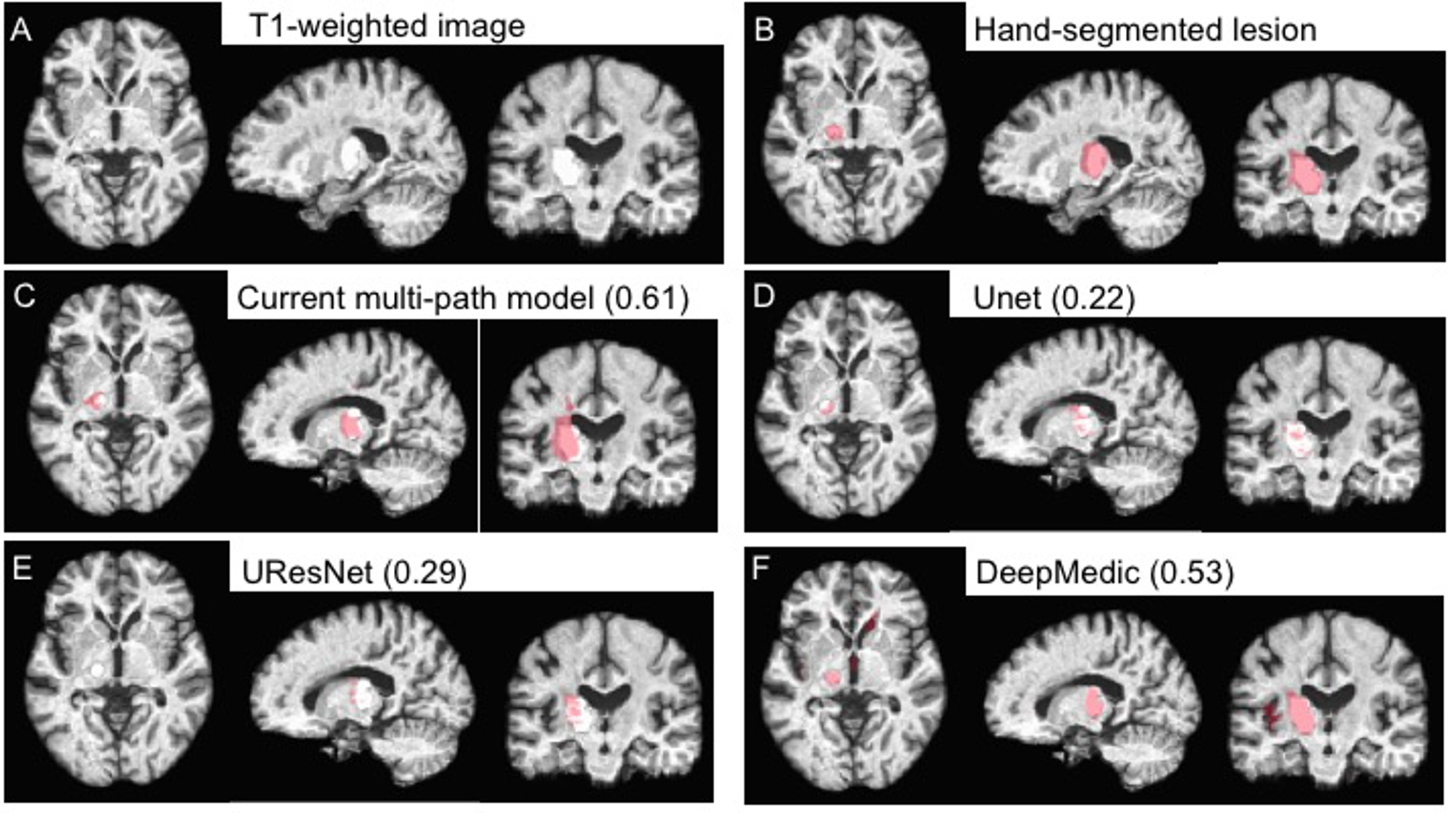}
\caption{Example of a relatively large (10,739 mm$^{3}$) lesion (A) along with its hand-segmented mask (B). The remaining panels show the lesion masks derived from the 5-fold cross-validation with all 99 scans for our 2.5D model (C) and the other CNN-based approaches (D-F). The label for each model is followed by the corresponding Dice value for the lesion mask it produced in parentheses. Lesion masks overlaid in red are rendered semi-transparent to visualize the overlap between the lesion and the mask. \label{large_lesion}}
\end{figure}

In all three cases, our method does very well on large lesions. In fact, when we train on KES+MCW and predict on ATLAS, the median Dice is above 0.8 for large lesions. In the cross-validation on all data combined, our model is significantly better than all methods except for DeepMedic, with p-values below 0.05. An example of a larger lesion is shown in Figure~\ref{large_lesion}. The output lesion masks in red show our method and the other three to be qualitatively similar. An apparent exception is DeepMedic, which misidentifies tissue in the right-hemisphere as being lesioned. This mis-identification would seem to be an exception, however, given the similar numeric performance between our method and DeepMedic.

\begin{figure}[h]
\includegraphics[scale=.93]{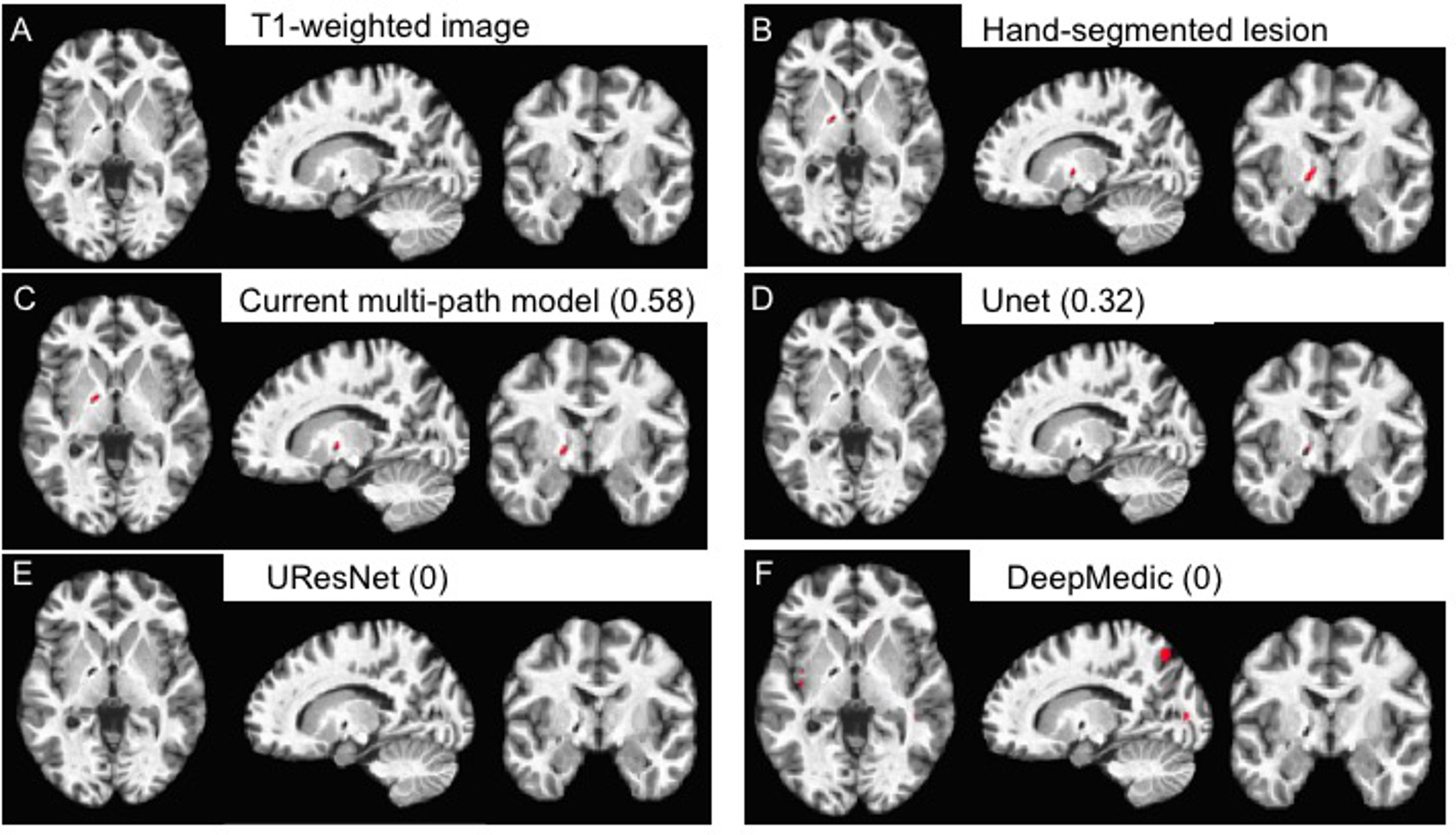}
\caption{Example of a relatively small (85 mm$^{3}$) lesion (A) along with its hand-segmented mask (B). The remaining panels show the lesion masks derived from the 5-fold cross-validation with all 99 scans for our 2.5D model (C) and the other CNN-based approaches (D-F). The label for each model is followed by the corresponding Dice value for the lesion mask it produced in parentheses. Lesion masks are overlaid in red. Note that the lesion masks derived from the DeepMedic model (F) are valse positives rather than actual lesions. \label{small_lesion}}
\end{figure}

Smaller lesions, on the other hand, are associated with lower median Dice values overall, as generally expected. DeepMedic has particular difficulty with smaller lesions, whereas our system shows significantly greater accuracy than DeepMedic and UResNet. Interestingly, the distribution of Dice values for small lesions clusters towards the high end in the cross-validation setup with the most training data (all three datasets combined). This suggests that still more data would enable the model to achieve better accuracy at identifying small lesions. An example of a smaller lesion classification for the combined data cross-validation scenario is shown in Figure~\ref{small_lesion}. This figure shows how the similarity of the overall contours of the model-based lesion masks (C-F) match up with the hand-segmented lesion mask (B). It also illustrates the face validity of the Dice coefficient, where higher Dice values also qualitatively correspond better to the hand-segmented lesion mask.

CNNs are a type of neural network, and what neural networks learn depends on what information is in the training data \cite{plaut1996understanding}. In the cross-study scenario where we train on KES+MCW and test on ATLAS, the distribution of Dice values for smaller lesions is spread somewhat uniformly. However, when the network is trained on the ATLAS data and tested on the KES+MCW set, performance is worse. Thus the general rule that the information in the training dataset largely determines what the model can learn is also shown here for detecting small lesions.

\subsection{Consolidating multi-path outputs}
\begin{figure}[h]
\centering
\includegraphics[scale=.4]{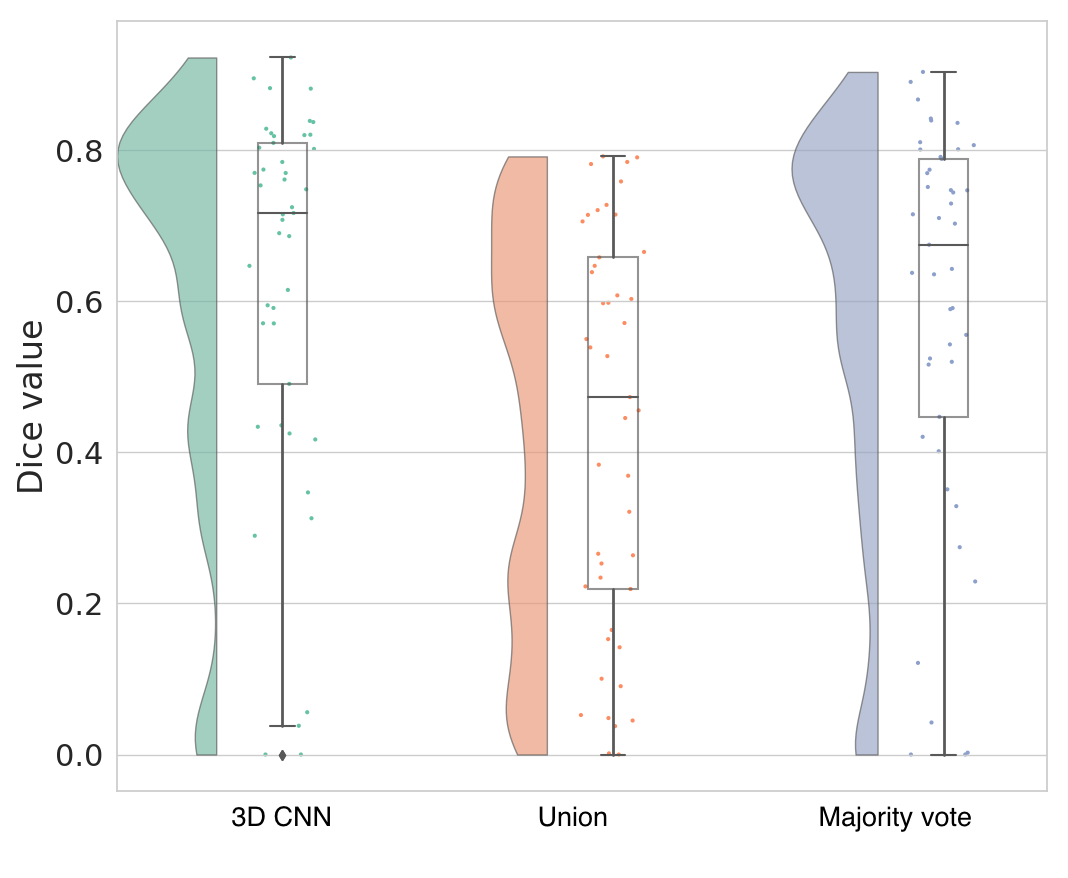} \\
\centering
\footnotesize
\begin{tabular}{cccccccc}\hline
3D CNN & Union & Majority vote \\ 
0.62 & 0.43 & 0.58 \\  \hline
\end{tabular}
\caption{Raincloud plot of Dice coefficient values of three different post-processing approaches in our system as given by five-fold cross validation on KES+MCW images combined. Mean Dice values for each approach are presented in the accompanying table. \label{postprocess}}
\end{figure}

Previous multi-path approaches use a majority vote to combine outputs from different paths \cite{lyksborg2015ensemble}. We compare our 3D CNN for combining multi-path outputs to using the majority vote and a simple union. In the union method, if at least one pixel has a one across the paths then the aggregated output also has a one in that pixel. Figure~\ref{postprocess} shows that the union clearly performs more poorly than majority vote and our 3D CNN. Between the two better performing methods, the 3D CNN is reliably better than majority vote by a 4\% margin with a p-value of 0.004. 
Also compared to post-processing with majority vote, the Dice values of the 3D CNN are concentrated more towards the high end.

\subsection{Mulit-modal T1 vs. T1+FLAIR}
Our basic U-Net model is multimodal (specifically, bimodal) in that it allows for different image formats. Since the current project is focused exclusively on left hemisphere lesions, we present the model with T1 and FLAIR image formats of the lesioned left hemisphere. Below in Table~\ref{kcw2} we show the cross-validation accuracy of our model on the KES and MCW images. When presented together, there is no significant difference between the two. However, if we look at just KES images that contain smaller lesions (and more recent, in the less than 5-week post-stroke range), then adding FLAIR confers a significant advantage. In the case of MCW images only that have lesions exclusively in the chronic epoch (at least 6 months post-stroke), the T1 images alone actually result in better performance than when the corresponding FLAIR images are added. This pattern corresponds with the standard clinical observation that FLAIR scans are useful for more recent stroke lesions but less so for those in the chronic phase \cite{ricci1999comparison}. Such correspondence lends additional face validity to our model.

\begin{table}[!h]
\centering
\footnotesize
\begin{tabular}{cccccccc}\hline
Data & T1 & T1+FLAIR & Wilcoxon rank test p-value & Average lesion size (in pixels) \\ 
KES+MCW & 0.59 & 0.63 & 0.2 & 58388 \\  
KES & 0.47 & 0.58 & 0.004$^*$ & 34054 \\  
MCW & 0.74 & 0.68 & 0.002$^*$ & 88804 \\  \hline
\end{tabular}
\caption{Mean Dice coefficients of our method on T1 vs. T1+FLAIR images on Kessler+MCW. Also shown are Wilcoxon rank test p-values and average lesion size of images in the combined and individual datasets 
\label{kcw2}}
\end{table}


\section{Discussion}

Here we have created, trained, and tested a new multi-path 2.5D convolutional neural network. The fractional designation on the dimension comes from its use of nine different 2D paths, followed by concatenation of the learned features across the paths, which are then passed to a 3D CNN for post-processing. This 2.5D design combines flexible and efficient 2D paths that process the data in different canonical orientations and normalizations with a 3D CNN that combines the 2D features in a way that informs the final 3D image output. Comparison of our system to previous efforts shows that CNN-based systems outperform more traditional machine-learning approaches based on random forests or Gaussian naive Bayes algorithms. Compared to other CNN systems, our system shows reliably superior performance in its ability to automatically segment stroke lesions from healthy tissue in the left hemisphere.

As methods such as this continue to improve the automated segmentation of brain lesions, a question arises. How good is good enough? An intuitive answer to this question comes from human expert raters. As mentioned in the Introduction, human expert raters have been shown to produce lesion segmentations with overlapping volumes between raters in the 67\% to 78\% range \cite{neumann2009interrater,fiez2000,liew2018large}, though 73\% may be a more realistic upper value given the highly expert raters and limited scope of the data used by Neumann et al. \cite{neumann2009interrater} to obtain the 78\% value. The Dice coefficient used here is a formal measure of degree of spatial overlap that ranges between 0 and 1. Therefore a Dice coefficient in the 0.67 range can be considered to be at the edge of the human expert gold standard. When combining the datasets and performing iterative training and testing using standard 5-fold cross-validation, the lesion traces from our model overlap with human experts with a mean Dice coefficient of 0.62. While the 0.67 to 0.73 human benchmark range should be interpreted with caution because those numbers are based on data that are not identical to the data considered here, the accuracy of our system relative to previous efforts does suggest that deep learning-based CNN methods are beginning to approach human expert level accuracy for stroke lesion segmentation.

\subsection{Future directions}
An alternative to our system is to have a multi-modal 3D U-Net instead of the current 2D ones. While promising, it may be difficult to implement in practice. Training a 3D CNN involves adjusting many more parameters than for a 2D CNN, and would therefore require more data to train. 
A second future direction is to extend our current left hemisphere-focused system to include lesions to the right hemisphere. This extension should be relatively straightforward, as nothing is preventing our current system from being trained and tested on images with lesions to either hemisphere.





\subsection{Conclusion}
We have presented a multi-path, multi-modal convolutional neural network system for identifying lesions in brain MRI images. While the data with which our model is trained and tested includes exclusively left hemisphere lesions, our model can be trained and tested on lesions present anywhere in the brain. In cross-study and cross-validation tests, our model shows superior performance compared to existing CNN and non-CNN based machine learning methods for lesion identification. Our method extends previous efforts showing relatively high segmentation accuracy for large lesions. Given sufficient data, it markedly improves on previous efforts by being able to segment smaller lesions as well. We provide freely available open source code to train and test our model.

This advance in performance is critically significant, as it brings the field closer to removing the bottleneck of having human experts spend numerous hours hand-segmenting brain lesions on MRI scans. Once automated methods are sufficiently accurate and widely available, they will free up researchers to focus their time on other critical aspects of neuropsychological data acquisition and analysis. The hope is this re-allocation of expert resources will help advance the pace at which we can further our understanding of the critical neural basis of thinking and behavior.

\section{References}

\bibliographystyle{unsrt}

\bibliography{my_bib}
\end{document}